\documentclass[sigconf]{acmart}

\ifdefined\pdfsuppressptexinfo
\fi

\setcopyright{none}
\acmConference[WWW '27]{The ACM Web Conference 2027}{May 10--14, 2027}{Dublin, Ireland}
\acmBooktitle{}
\acmYear{2027}
\copyrightyear{}
\acmDOI{}
\acmISBN{}

\usepackage{amsmath}
\usepackage{multirow}
\usepackage{colortbl}
\usepackage{tabularx}
\usepackage{makecell}
\usepackage{enumitem}
\usepackage{tcolorbox}

\definecolor{promptframeblue}{RGB}{25,50,120}
\definecolor{promptbgblue}{RGB}{235,240,255}
\newtcolorbox[auto counter]{ForecastPrompt}[1][]{
  float,
  title={Template~\thetcbcounter: Fast--Slow--Reflect Prompt},
  colframe=promptframeblue,
  colback=promptbgblue,
  colbacktitle=promptframeblue,
  coltitle=white,
  #1
}

\newcommand{\model}{CastFSR}

\begin{document}

\title[CastFSR: Fast--Slow--Reflect Agentic Reasoning for Context-Aware Time Series Forecasting]{CastFSR: Fast--Slow--Reflect Agentic Reasoning for Context-Aware Time Series Forecasting}
\author{Xiaoyu Tao}
\affiliation{%
  \institution{State Key Laboratory of Cognitive Intelligence, University of Science and Technology of China}
  \city{Hefei}
  \state{Anhui Province}
  \country{China}
}
\email{txytiny@mail.ustc.edu.cn}

\author{Mingyue Cheng}
\affiliation{%
  \institution{State Key Laboratory of Cognitive Intelligence, University of Science and Technology of China}
  \city{Hefei}
  \state{Anhui Province}
  \country{China}
}
\email{mycheng@ustc.edu.cn}

\author{Bokai Pan}
\affiliation{%
  \institution{State Key Laboratory of Cognitive Intelligence, University of Science and Technology of China}
  \city{Hefei}
  \state{Anhui Province}
  \country{China}
}
\email{bk2585934928@mail.ustc.edu.cn}

\author{Ze Guo}
\affiliation{%
  \institution{State Key Laboratory of Cognitive Intelligence, University of Science and Technology of China}
  \city{Hefei}
  \state{Anhui Province}
  \country{China}
}
\email{gz1504921411@mail.ustc.edu.cn}

\author{Hao Xue}
\affiliation{%
  \department{Thrust of Urban Governance and Design}
  \institution{The Hong Kong University of Science and Technology (Guangzhou)}
  \city{Guangzhou}
  \state{Guangdong}
  \country{China}
}
\email{haoxue@hkust-gz.edu.cn}
\renewcommand{\shortauthors}{Tao et al.}

\begin{abstract}
Time series forecasting is a fundamental task in temporal data mining, with applications ranging from web service monitoring to energy systems. Numerical forecasting models mainly capture historical patterns, but changing contextual factors can affect future dynamics. A key challenge is determining when and how contextual evidence should guide revisions to numerical forecasts and assessing whether these revisions are consistent with temporal patterns and domain knowledge. To address this challenge, we propose CastFSR, a fast--slow--reflect agentic framework for context-aware time series forecasting. During fast-thinking forecasting, CastFSR analyzes numerical time-series features and uses the resulting diagnostic evidence to select a numerical forecaster and construct a data-driven forecast prior. Building on this prior, slow deliberative reasoning identifies relevant contextual factors and uses the evidence extracted from these factors to guide selective forecast refinement. Reflective evaluation then reviews the revised forecast against temporal patterns, contextual evidence, and task-specific domain constraints, providing feedback for further correction when needed. A shared state retains numerical priors, contextual evidence, intermediate forecasts, and review feedback to guide forecasting decisions across the three stages. The framework supports both training-free coordination with general-purpose LLMs and two-stage post-training for compact LLM coordinators. Workflow-aware trajectory distillation (WATD) uses supervised fine-tuning on teacher trajectories to learn state-conditioned multi-turn tool use and forecasting decisions, while trajectory-level policy optimization (TLPO) further optimizes the coordinator policy through reinforcement learning guided by forecasting accuracy and workflow feedback. Experiments on multiple real-world datasets, conducted under an offline evaluation protocol using historical context, demonstrate competitive performance against \mbox{representative baselines}.
\footnote{Our code is available at \url{https://github.com/Xiaoyu-Tao/CastFSR}.}

\end{abstract}

\ccsdesc[500]{Computing methodologies~Artificial intelligence}
\keywords{Context-aware time series forecasting, Agentic reasoning,   Trajectory distillation, Reinforcement learning}

\maketitle

\section{Introduction}

Time series forecasting (TSF) supports operational decisions in applications ranging from IT service monitoring~\cite{palaskar2024automixer} to energy systems~\cite{lago2021forecasting}. Historical observations reveal trends, seasonality, and temporal dependencies~\cite{box2015time,cheng2025convtimenet}, but changing contextual factors, such as calendar effects, weather conditions, and external events, can alter future dynamics~\cite{williams2024context,wang2024timexer}. Context-aware TSF therefore draws on both numerical observations and contextual information~\cite{kamarthi2022camul}. The challenge is to determine how contextual factors should influence predictions while keeping predictions consistent with numerical evidence, domain knowledge, and \mbox{task-specific constraints}.

Numerical forecasting has progressed from statistical models~\cite{box2015time} and neural forecasters~\cite{cheng2025convtimenet,Yuqietal-2023-PatchTST} to pretrained time series foundation models~\cite{ansari2025chronos,woo2024unified}. Approaches based on large language models (LLMs) introduce language representations~\cite{jintime,xue2023promptcast} and explicit temporal reasoning~\cite{cheng2025can}, while context-aware methods incorporate exogenous variables~\cite{wang2024timexer}, contextual text~\cite{wu2026aurora}, and event information~\cite{wang2026vot}. Agentic frameworks further support tool interaction and evidence acquisition~\cite{pan2026castflow,zhao2025timeseriesscientist,zhang2026search}, expert selection~\cite{yan2026moiraiagent}, and iterative forecast refinement~\cite{pan2026castflow,zhang2025alphacast}. These studies provide complementary capabilities, but combining them does not by itself ensure that forecast revisions are supported by numerical priors and contextual evidence. Subsequent review must assess whether departures from historical patterns are justified and consistent with domain knowledge. The central challenge is therefore to keep forecasting decisions connected to their supporting rationale throughout numerical prediction, contextual reasoning, and \mbox{forecast review}.

\begin{figure}[!t]
    \centering
    \includegraphics[width=0.95\columnwidth]{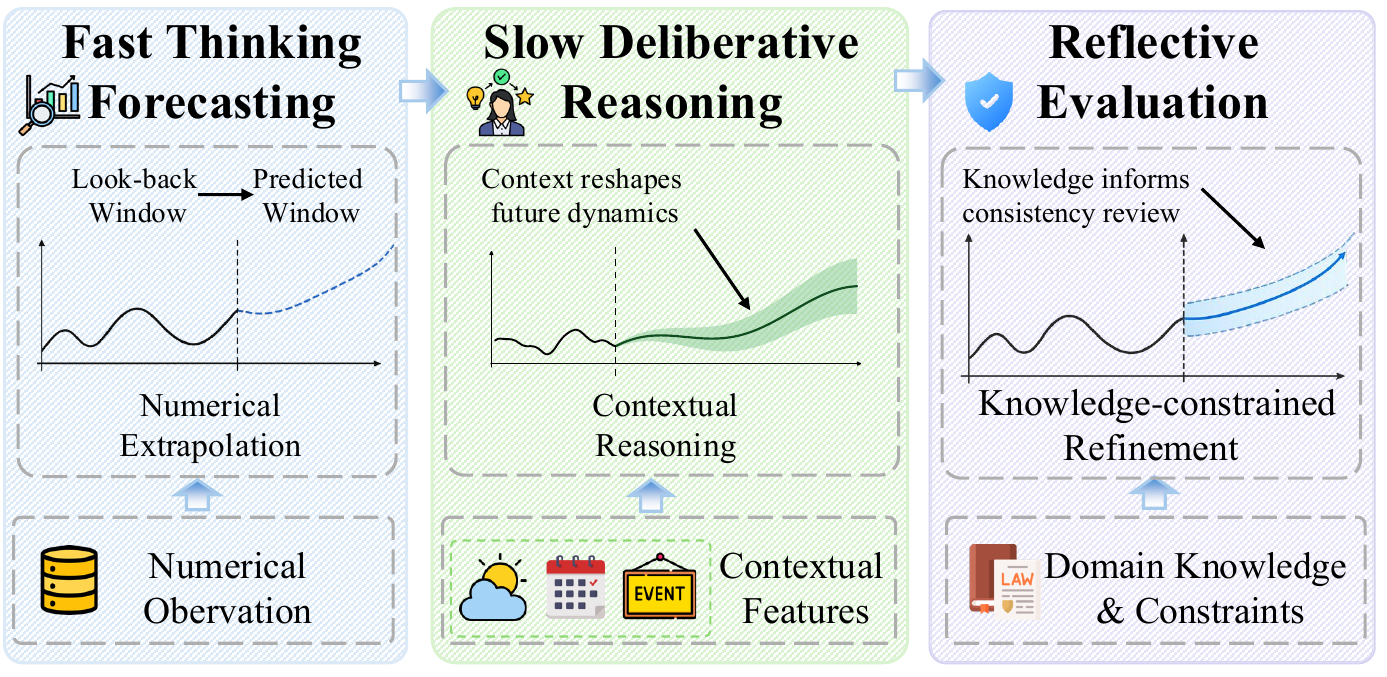}
    \caption{Conceptual illustration of the three interdependent forecasting processes in \model{}: numerical prior construction, contextual forecast refinement, and \mbox{knowledge-guided review}.}
    \Description{Three connected panels conceptually illustrate fast-thinking forecasting, slow deliberative reasoning, and reflective evaluation. Historical observations support numerical extrapolation, contextual features guide forecast refinement, and domain knowledge and constraints guide a review of the revised forecast.}
    \label{fig:motivation}
\end{figure}

Figure~\ref{fig:motivation} conceptually illustrates three interdependent forecasting processes and their complementary roles in constructing, refining, and reviewing forecasts. First, numerical temporal pattern modeling must establish a suitable forecast prior from diverse trends, seasonal patterns, and temporal dependencies. Different patterns may favor different numerical forecasters, making expert selection important for subsequent refinement~\cite{yan2026moiraiagent,zhao2025timeseriesscientist}. Second, contextual factor modeling must identify relevant contextual features~\cite{williams2024context,wang2026vot} and determine whether the historical evidence available at the forecast origin supports revisions within the forecast horizon. Third, knowledge-guided constraint checking must assess the revised forecast together with its supporting evidence against temporal patterns, domain knowledge, and task-specific constraints~\cite{zhang2025alphacast,pan2026castflow}. Contextual changes can justify departures from historical patterns, so this review should distinguish supported changes from unsupported adjustments rather than simply preserve \mbox{the prior}.

To address these challenges, we propose CastFSR, a fast--slow--reflect agentic framework for context-aware TSF. During fast-thinking forecasting, an LLM coordinator analyzes numerical time-series patterns using diagnostic tools and selects a numerical forecaster to construct a data-driven forecast prior. Slow deliberative reasoning then identifies relevant factors from historical contextual evidence and reasons about their potential effects on future dynamics to selectively refine the prior. Reflective evaluation reviews the revised forecast against temporal patterns, contextual evidence, and domain-specific constraints, providing feedback for iterative correction. A shared state connects numerical priors, contextual evidence, and candidate forecasts across the three stages, retaining the available evidence and rationale for \mbox{forecast review}. CastFSR-Zero uses an off-the-shelf LLM without task-specific coordinator training, while CastFSR-R1 learns the fast--slow--reflect coordination policy using a compact 4B model. Training covers diagnostic and forecaster selection, contextual refinement, and reflective review within the same workflow, extending supervision beyond final numerical outputs. Workflow-aware trajectory distillation (WATD) uses supervised fine-tuning  on teacher trajectories to learn forecasting decisions conditioned on the evidence available at each step. Trajectory-level policy optimization (TLPO) then optimizes the coordination policy through reinforcement learning, guided by forecasting accuracy and workflow feedback. Experiments on multiple real-world datasets under an offline historical-context evaluation protocol demonstrate the competitive forecasting performance of CastFSR-R1 across diverse forecasting scenarios.

Our main contributions are summarized as follows:

\begin{itemize}
    \item We propose CastFSR, a fast--slow--reflect agentic framework that coordinates forecasting decisions through a structured shared state, with revisions guided by numerical priors and contextual evidence.

    \item We train a compact 4B coordinator to learn the fast--slow--reflect coordination policy through WATD and TLPO, combining state-conditioned trajectory supervision with feedback from complete \mbox{forecasting trajectories}.

    \item We demonstrate competitive performance on multiple real-world datasets under an offline evaluation protocol using historical context, with ablation studies assessing workflow components and \mbox{post-training strategies}.
\end{itemize}

\section{Related Work}
\subsection{Numerical Time Series Forecasting}
Numerical forecasting spans statistical modeling, neural representations, and pretraining~\cite{cheng2025comprehensive}. Statistical models such as exponential smoothing~\cite{hyndman2008automatic} and ARIMA~\cite{box2015time} capture trend, seasonality, and autoregression. Neural forecasters use convolutions~\cite{cheng2025convtimenet}, patch-based modeling~\cite{Yuqietal-2023-PatchTST}, and cross-variable attention~\cite{liuitransformer}, while foundation models such as Moirai~\cite{woo2024unified} and Chronos-2~\cite{ansari2025chronos} enable transfer through pretraining. DynaME~\cite{hong2026dyname} handles recurring and emergent concept drift through periodic specialists and a stable general expert for online forecasting. For contextual modeling, TimeXer~\cite{wang2024timexer} models exogenous-to-endogenous dependencies, and Chronos-2 incorporates covariates. Despite these advances, most methods emphasize numerical patterns rather than explicit reasoning over contextual evidence~\cite{ma2026quizsf}. TraffiDent~\cite{gou2025traffident} aligns traffic observations with incident records to support post-incident forecasting, highlighting the relevance of external events.

\begin{figure*}[t]
    \centering
    \includegraphics[width=0.98\linewidth]{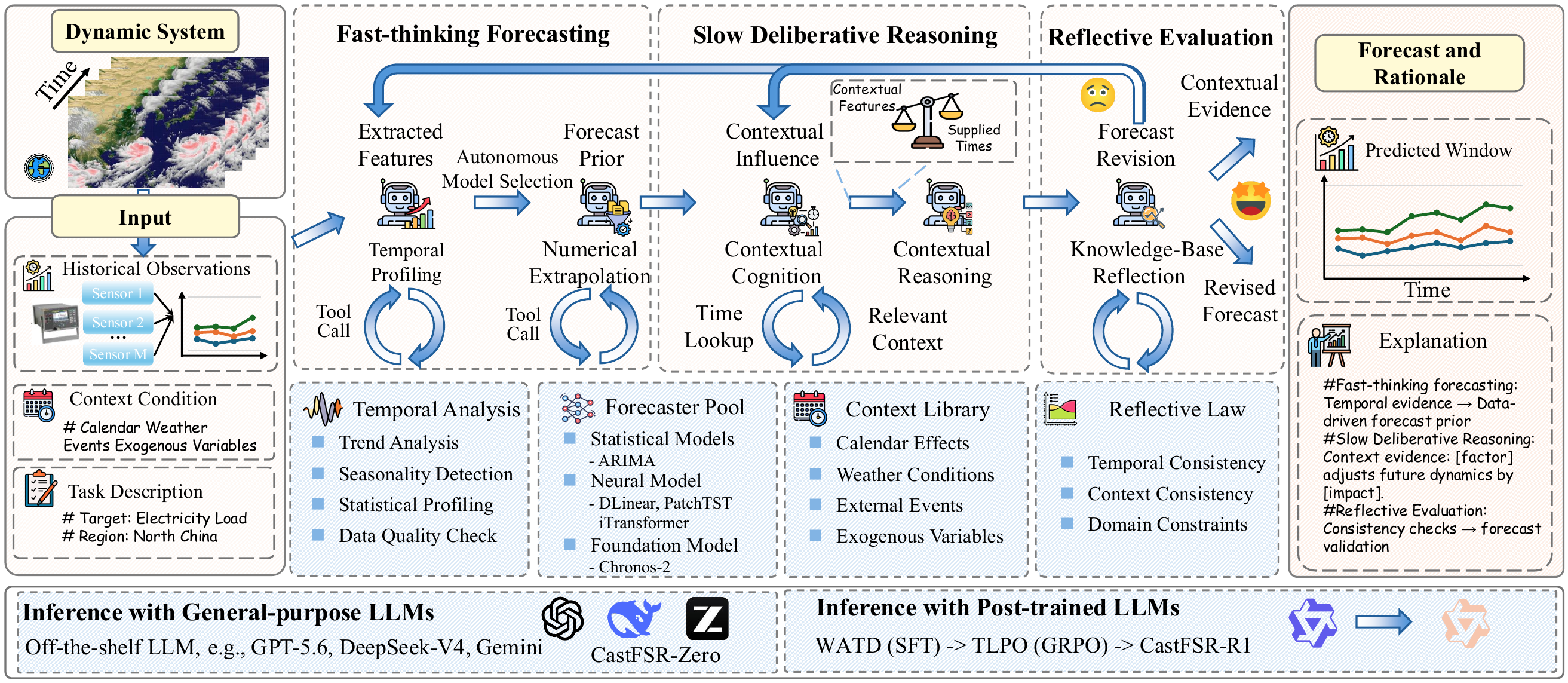}
    \caption{Overview of CastFSR: a fast--slow--reflect workflow for context-aware forecasting with either a training-free LLM coordinator or a post-trained compact coordinator.}
    \Description{Historical observations and contextual evidence support fast-thinking forecasting, followed by separate generations for slow deliberative reasoning and reflective evaluation. The forecast prior is refined and reviewed for temporal, contextual, and domain consistency. General-purpose and post-trained LLMs execute the workflow.}
    \label{fig:framework}
\end{figure*}

\subsection{LLM-based Time Series Forecasting}

LLM forecasting spans representation learning, reasoning, and agentic coordination. PromptCast~\cite{xue2023promptcast} and Time-LLM~\cite{jintime} use prompting and reprogramming, respectively; TEMPO~\cite{cao2024tempo} combines temporal decomposition with prompt tuning. GPT4MTS~\cite{jia2024gpt4mts}, TokenCast~\cite{tao2025values}, and ChatTime~\cite{wang2025chattime} integrate numerical and textual inputs; ExoLLM~\cite{huang2025exollm} models exogenous information through prompts and cross-modal attention. TimeReasoner~\cite{cheng2025can} studies inference-time forecast reasoning, while Time-R1 learns it through post-training. VeriTime~\cite{zhou2026veritime} develops time-series reasoning through process-verifiable data synthesis, data scheduling, and reinforcement learning. MemCast~\cite{tao2026memcast} uses memory-conditioned reasoning and refinement, and TimeOmni-1~\cite{guan2026timeomni} examines reasoning across time-series tasks. TimeSeriesScientist~\cite{zhao2025timeseriesscientist} combines diagnostics, model selection, and ensembles. Recent work further emphasizes explicit evidence acquisition, iterative refinement, and flexible coordination across forecasting stages. AlphaCast~\cite{zhang2025alphacast} integrates case-based numerical references, contextual refinement, and knowledge-guided reflection in an investigator--generator--reflector workflow. CastFlow~\cite{pan2026castflow} combines evidence acquisition, prior-guided refinement, and reflective checks, with post-training focused on its specialized forecasting module. MoiraiAgent~\cite{yan2026moiraiagent} combines expert selection and contextual adjustment, while CastClaw~\cite{tao2026castclaw} supports human-guided forecasting through tool orchestration.

\subsection{Agentic Post-Training Methods}
Agentic post-training builds on demonstration learning, reward-based policy optimization, and distillation. InstructGPT~\cite{ouyang2022instructgpt} combines supervised fine-tuning (SFT) on demonstrations with reinforcement learning from human feedback (RLHF), using reward models learned from human preferences. AgentTuning~\cite{zeng2024agenttuning} extends demonstration learning to agent interaction trajectories. Reinforcement learning with verifiable rewards (RLVR) uses automatically checkable outcomes, as in the reasoning-oriented RL of DeepSeek-R1~\cite{guo2025deepseekr1}. GRPO, introduced in DeepSeekMath~\cite{shao2024deepseekmath}, estimates relative advantages within groups of sampled responses without a separate critic. For multi-turn agents, RAGEN~\cite{wang2025ragen} studies trajectory-level optimization, while StepPO~\cite{wang2026steppo} aligns credit assignment and policy updates with complete interaction steps. Agent-R1~\cite{cheng2025agentr1} provides modular training with step-level trajectory representations and flexible context management. In forecasting, Cast-R1~\cite{tao2026castr1} combines teacher-trajectory SFT with memory-supported multi-turn GRPO for diagnostic tool use, forecaster routing, and forecast refinement.

CastFSR links selective prior revision and knowledge-guided review through shared evidence and rationale. Shared state connects diagnostic-guided forecaster selection, numerical priors, historical context, and corrective feedback within a fast--slow--reflect workflow. The same workflow supports training-free LLM coordination and a compact coordinator trained on state-conditioned multi-turn decisions through trajectory supervision and feedback on forecasting accuracy and workflow execution.

\section{The Proposed CastFSR}

\subsection{Problem Formulation}
\label{subsec:problem_formulation}
Given historical target observations, historical contextual features, and task description and domain metadata at the forecast origin, context-aware time series forecasting aims to predict future target values by incorporating temporal patterns, contextual evidence, and domain knowledge. Here, contextual evidence refers to historical non-target information that may help interpret observed dynamics and inform forecasting decisions. For example, in web workload forecasting, historical request volumes serve as target observations, while service telemetry provides potential contextual evidence. Formally, the forecasting task is defined as:
\begin{equation}
\hat{\mathbf Y}_{1:H} =
f(\mathbf X_{1:L}, C, D),
\label{eq:forecasting_task}
\end{equation}
where $\mathbf X_{1:L}$ denotes the target history of length $L$, $C$ contains historical contextual features, $D$ contains the task description and domain metadata, $\hat{\mathbf Y}_{1:H}$ denotes the $H$-step forecast, and $f$ is the forecasting function evaluated at origin $t_0$. Only information available at the forecast origin is used, excluding forecast-horizon target and covariate values. 

\subsection{Framework Overview}
\label{subsec:overview}
As shown in Figure~\ref{fig:framework}, \model{} organizes context-aware forecasting into a fast--slow--reflect workflow driven by an LLM coordinator. Given target history, historical context, and task metadata, fast-thinking forecasting selects a diagnostic tool and a numerical forecaster to construct a data-driven prior. Slow deliberative reasoning interprets contextual evidence alongside the prior and diagnostics to determine whether and how the forecast should be revised. Reflective evaluation then reviews the candidate and its rationale against temporal patterns, contextual evidence, and domain constraints, providing feedback for correction as needed.

A forecasting harness connects these stages through shared state that retains diagnostic evidence, the numerical prior, the candidate forecast, supporting rationale, and review feedback. It manages tool execution, stage transitions, output checks, and retry budgets, while the coordinator makes decisions from the available evidence. The same workflow supports two instantiations: \model{}-Zero uses a general-purpose LLM without task-specific coordinator training; \model{}-R1 uses a compact coordinator adapted through WATD followed by TLPO. In both cases, numerical experts retain their own fitting and inference procedures.

\subsection{Fast-Slow-Reflect Forecasting}
\label{subsec:harness}
The harness combines shared state and tool interfaces with the fast--slow--reflect workflow. It connects numerical evidence, contextual reasoning, and forecasting decisions by maintaining accumulated state and managing the execution protocol.

\subsubsection{\textbf{Shared State and Tool Interfaces}}
\label{subsec:shared_state}

The shared state at each interaction step is:
\begin{equation}
s_k=(D,\mathbf X_{1:L},C,F_k,
     \hat{\mathbf Y}^{\rm prior}_k,\hat{\mathbf Y}^{\rm slow}_k,M_k),
\label{eq:castfsr_state}
\end{equation}
where $F_k$ is diagnostic evidence, $\hat{\mathbf Y}^{\rm prior}_k$ is the numerical forecast prior, and $\hat{\mathbf Y}^{\rm slow}_k$ is the candidate produced by the slow stage. $M_k$ records tool interactions, routing, execution, and available reasoning for the current forecasting instance. Diagnostic and prediction fields are initially empty. The fast stage writes diagnostic evidence and the forecast prior. The slow stage adds its candidate and supporting rationale. Reflect reads these together with historical context to assess proposed revisions, and its feedback accompanies the candidate in subsequent coordinator requests.
Diagnostic interfaces return evidence about temporal characteristics, while forecasting interfaces invoke numerical experts and return forecast priors. The harness incorporates these tool observations into shared state for subsequent decisions. Each coordinator request combines a system prompt, stage instructions, and the current state snapshot. The situation and covariate block is cached within a run. Contextual fields are aligned with historical observation timestamps, and all LLM reasoning uses values in their original units. Z-score standardization is reserved for evaluation. These state-conditioned interactions also form the trajectories used for coordinator post-training.

\subsubsection{\textbf{Fast--Slow--Reflect Workflow}}
\label{subsec:fsr_forecasting}
\label{subsec:fast_forecasting}
\label{subsec:slow_deliberation}
\label{subsec:reflective_evaluation}
The workflow specifies how the coordinator reads and updates shared state across the fast, slow, and reflect stages. Stage transitions, output checks, and retry budgets are specified by the harness. Within this protocol, the coordinator selects diagnostics and forecasters, generates contextual refinements, and reviews candidate forecasts.

\paragraph{\textbf{Fast-thinking Forecasting}.}
The fast stage analyzes temporal characteristics and adaptively selects a numerical forecaster to construct a data-driven forecast prior. Differences in trend, seasonality, dependence, volatility, and data quality motivate diagnostic-guided selection. Available diagnostics summarize basic statistics, temporal dependence, residual behavior, data quality, and temporal events.
The default protocol selects one diagnostic tool, obtains its result, and updates the shared state before selecting one \mbox{numerical forecaster}:
\begin{align}
q&\sim\pi_\theta(\cdot\mid s_0),
    \qquad F=q(\mathbf X_{1:L}),\label{eq:fast_diagnostic}\\
m&\sim\pi_\theta(\cdot\mid s_1),\label{eq:fast_expert}
\end{align}
where $Q$ and $P$ are the diagnostic tool set and numerical forecaster pool, $\pi_\theta$ is the coordinator policy with parameters $\theta$, $q\in Q$ is the selected diagnostic tool, $F$ is its output, and $m\in P$ is the selected forecaster. The initial state $s_0$ becomes $s_1$ after incorporating $F$. The pool comprises ARIMA, DLinear, PatchTST, iTransformer, and Chronos-2. The selected expert produces the horizon-length prior:
\begin{equation}
\hat{\mathbf Y}^{\rm prior}=T_{\rm pred}(\mathbf X_{1:L},C,m),
\label{eq:forecast_prior}
\end{equation}
where $T_{\rm pred}$ is the forecasting interface for the selected expert $m$, and $\hat{\mathbf Y}^{\rm prior}$ is its $H$-step numerical forecast. The harness writes the diagnostic evidence, selected expert identity, and forecast prior into shared state. Selection is conditioned on visible evidence and tool descriptions. Future target values are unavailable. All evaluated expert calls, including Chronos-2, use historical target observations and historical covariates according to their input interfaces. The prior can therefore already incorporate historical context, providing a numerical reference for the slow stage.

\paragraph{\textbf{Slow Deliberative Reasoning}.}
The slow stage interprets historical contextual evidence to guide selective forecast refinement. Although the numerical prior may already incorporate historical covariates, this stage explicitly reasons about whether and how the available evidence supports revising that prior. Its prompt requests situation analysis to identify relevant contextual factors and their relation to observed target behavior, followed by forecast reasoning about the direction and timing of supported changes within the horizon. These judgments are grounded in historical observations, diagnostic evidence, and domain information.

In the evaluated setting, the Context Library in Figure~\ref{fig:framework} denotes each dataset's existing contextual covariates organized by observation timestamp. It is constructed from the non-target fields accompanying the time series. At each forecast origin, retrieval follows the look-back observation timestamps to obtain $C$, excluding forecast-horizon covariates. For a different forecast origin, $C$ is taken from the corresponding historical slice; the situation and covariate block is cached within a run. Library contents are domain-specific, such as weather variables for wind power and load and generation variables for electricity prices, while timestamp-based access and state integration are shared across tasks.

Supported revisions can include local changes or sustained level and trend adjustments. The coordinator jointly generates a rationale and a \mbox{candidate forecast}:
\begingroup
\postdisplaypenalty=10000
\begin{equation}
(R_{\rm slow},\hat{\mathbf Y}^{\rm slow})\sim\pi_\theta
 (\cdot\mid\mathbf X_{1:L},F,
 \hat{\mathbf Y}^{\rm prior},C,D,M),
\label{eq:slow_candidate}
\end{equation}
\endgroup
where $R_{\rm slow}$ explains the proposed revisions, $\hat{\mathbf Y}^{\rm slow}$ is the candidate forecast, and $M$ contains available execution records. The prompt encourages preserving useful prior structure wherever contextual evidence is insufficient to justify a change. Selective refinement is a prompt-guided decision expressed through the generated candidate: $\hat y_h^{\rm slow}-\hat y_h^{\rm prior}$ describes the resulting adjustment. The coordinator generates forecast values directly, with no separate residual action, magnitude bound, adaptive retrieval-window action, or estimated causal-effect model. The harness retains the candidate and its available reasoning and passes them, together with the accumulated evidence, to the \mbox{reflect stage}.

\paragraph{\textbf{Reflective Evaluation}.}
The reflect stage evaluates the revised forecast against temporal, contextual, and domain-specific constraints. A separate coordinator generation receives the slow-stage candidate and its supporting evidence for knowledge-guided constraint checking. The review covers \emph{temporal consistency} with historical dynamics and the observation-window boundary; \emph{contextual consistency} of proposed revisions with retained evidence and rationale; and \emph{domain consistency} with task-specific knowledge and physical constraints in original target units.
Detected violations are passed, together with the candidate and retained evidence, to the next coordinator generation for correction and rechecking. The harness carries feedback between attempts and enforces at most three correction retries, producing the reviewed forecast $\hat{\mathbf Y}$. These checks address the specified rules without certifying semantic correctness.
The harness separately checks output validity, requiring an \texttt{<answer>} block with $H$ timestamped rows and finite values in plain decimal notation. Invalid format, count, or timestamps can trigger another attempt within budget. A valid final answer from this process terminates execution. Budget exhaustion is a policy failure, with evaluation fallbacks handled separately. An optional deterministic guard checks only the numerical prior.

\subsection{Model-Agnostic Agent Instantiations}
\label{subsec:post_training}
The harness is driven by an LLM coordinator that conditions each forecasting decision on the current shared state. Model-agnostic refers to the coordinator backbone: general-purpose LLMs and post-trained compact models execute the same fast--slow--reflect workflow through shared state and tool interfaces. We instantiate these two approaches as \model{}-Zero and \mbox{\model{}-R1, respectively.}

\subsubsection{Training-Free LLM Coordination}
\label{subsec:training_free_coordination}

\model{}-Zero supports inference with a general-purpose LLM. System prompts, tool descriptions, stage instructions, and shared state guide its multi-turn forecasting decisions. Recorded tool observations and available coordinator responses form teacher trajectories for subsequent distillation. Training-free refers to the coordinator, whose parameters remain unchanged. Numerical forecasting experts retain their respective fitting or pretraining procedures. This backbone flexibility allows the framework to incorporate advances in general-purpose LLM reasoning and tool use while retaining the shared workflow and tool interfaces.

\subsubsection{Workflow-Aware Agentic Post-Training}
\label{subsec:workflow_post_training}
\model{}-R1 adapts Qwen3-4B to drive the same harness through WATD followed by TLPO. Both stages update only the coordinator, retaining the workflow, tool interfaces, and numerical experts' fitting and inference procedures. The compact coordinator also offers a path to self-hosted inference for privacy-sensitive applications: target histories and contextual inputs can remain within a controlled environment when both the coordinator and tool services run locally.

Workflow-Aware Trajectory Distillation (WATD).
WATD uses supervised fine-tuning (SFT) to learn multi-turn forecasting decisions from \model{}-Zero trajectories. It preserves the evidence available at each teacher decision by minimizing:
\begin{equation}
L_{\rm WATD}(\theta)
 =-\mathrm E_{\tau\sim D_{\rm teacher}}\!\left[\sum_k\log\pi_\theta(a_k\mid s_k)\right],
\label{eq:watd_objective}
\end{equation}
where $D_{\rm teacher}$ is the teacher trajectory dataset, $\tau$ is a trajectory sampled from it, and $a_k$ is the teacher response paired with the preceding visible state $s_k$ at interaction step $k$. Diagnostic-selection responses condition on historical observations and task metadata, while forecaster-selection responses additionally use the returned diagnostic evidence. Contextual-refinement responses condition on the numerical prior, historical context, diagnostic evidence, and execution records; reflective-review responses additionally receive the slow-stage candidate and its supporting rationale. Tool observations provide conditioning evidence, while teacher responses are the supervised targets.

Trajectory-Level Policy Optimization (TLPO).
TLPO applies the GRPO algorithm~\cite{shao2024deepseekmath} to multi-turn trajectories sampled by the WATD-initialized coordinator within the harness. The core forecast reward measures accuracy against offline training targets, subject to output-validity checks. Step rewards combine this accuracy feedback with workflow feedback, including successful tool progress, output-format validity, forecast-length and timestamp checks, protocol violations, and unsuccessful completion, as detailed in Appendix~\ref{app:reward}.
With reward-side KL disabled, step rewards form the trajectory return. For a group of trajectories sampled from the same prompt, the return and \mbox{normalized advantage are}:
\begin{equation}
R(\tau_i)=\sum_k r_{ik},\qquad
\widehat A_i=\frac{R(\tau_i)-\mu_R}{\sigma_R+\epsilon},
\label{eq:tlpo_advantage}
\end{equation}
where $\tau_i$ is the $i$-th sampled trajectory, $r_{ik}$ is its reward at step $k$, $R(\tau_i)$ is the trajectory return, and $\widehat A_i$ is its normalized advantage. The quantities $\mu_R$ and $\sigma_R$ are the group mean and sample standard deviation of trajectory returns, and $\epsilon=10^{-6}$. The trainer broadcasts this trajectory-level advantage to generated tokens for policy optimization, assigning credit at the trajectory level without independent causal attribution to individual decisions. 


\begin{table*}[t]
\centering
\caption{Overall forecasting performance in MSE and MAE. Lower values indicate better performance. The best results are bold and the second-best are underlined.}
\label{tab:main_results}
\footnotesize
\setlength{\tabcolsep}{0.8pt}
\renewcommand{\arraystretch}{1.12}
\begin{tabularx}{\textwidth}{@{}l*{20}{>{\raggedleft\arraybackslash}X}@{}}
\toprule
\multirow{3}{*}{\textbf{Models}} & \multicolumn{10}{c}{Long-term forecasting} & \multicolumn{10}{c}{Short-term forecasting} \\
\cmidrule(lr){2-11} \cmidrule(lr){12-21}
 & \multicolumn{2}{c}{ETTh1} & \multicolumn{2}{c}{ETTh2} & \multicolumn{2}{c}{ETTm1} & \multicolumn{2}{c}{ETTm2} & \multicolumn{2}{c}{Wind} & \multicolumn{2}{c}{BE} & \multicolumn{2}{c}{DE} & \multicolumn{2}{c}{FR} & \multicolumn{2}{c}{NP} & \multicolumn{2}{c}{PJM} \\
\cmidrule(lr){2-3} \cmidrule(lr){4-5} \cmidrule(lr){6-7} \cmidrule(lr){8-9} \cmidrule(lr){10-11} \cmidrule(lr){12-13} \cmidrule(lr){14-15} \cmidrule(lr){16-17} \cmidrule(lr){18-19} \cmidrule(lr){20-21}
 & MSE & MAE & MSE & MAE & MSE & MAE & MSE & MAE & MSE & MAE & MSE & MAE & MSE & MAE & MSE & MAE & MSE & MAE & MSE & MAE \\
\midrule
ARIMA & 0.156 & 0.291 & 0.356 & 0.460 & 0.142 & 0.301 & 0.173 & 0.379 & 3.682 & 1.361 & 1.973 & 0.645 & 1.519 & 0.722 & 6.369 & 1.075 & 0.727 & 0.471 & 0.991 & 0.691 \\
Prophet & 0.670 & 0.550 & 1.757 & 0.927 & 0.091 & 0.216 & 0.216 & 0.313 & 7.870 & 2.093 & 1.701 & 0.714 & 1.426 & 0.792 & 6.769 & 1.172 & 0.533 & 0.451 & 0.357 & 0.468 \\
PatchTST & 0.117 & 0.218 & 0.261 & 0.392 & 0.084 & \textbf{0.167} & 0.139 & \underline{0.244} & 2.522 & 1.198 & 1.164 & 0.495 & 0.845 & 0.520 & 5.335 & 0.674 & 0.385 & 0.398 & 0.180 & 0.334 \\
iTransformer & 0.091 & 0.213 & 0.264 & 0.387 & 0.083 & 0.175 & 0.139 & 0.255 & 2.616 & 1.196 & 1.465 & 0.685 & 1.112 & 0.727 & 5.448 & 1.026 & 0.453 & 0.462 & 0.357 & 0.448 \\
ConvTimeNet & 0.098 & 0.221 & 0.261 & 0.382 & 0.081 & 0.169 & 0.135 & 0.247 & 2.545 & 1.207 & 1.516 & 0.515 & 0.832 & 0.609 & 5.213 & 0.731 & 0.391 & 0.385 & 0.230 & 0.352 \\
TimeXer & 0.126 & 0.264 & 0.252 & 0.377 & 0.063 & 0.218 & 0.135 & 0.249 & 2.508 & 1.226 & 1.753 & 0.585 & 0.934 & 0.665 & 5.307 & 0.771 & 0.420 & 0.399 & 0.228 & 0.347 \\
DLinear & 0.122 & 0.268 & 0.252 & 0.386 & 0.079 & 0.254 & 0.138 & 0.253 & 2.440 & 1.149 & 1.677 & 0.600 & 1.005 & 0.687 & 5.203 & 0.791 & 0.488 & 0.446 & 0.269 & 0.367 \\
TimesFM & 0.116 & 0.266 & 0.295 & 0.405 & 0.086 & 0.241 & 0.200 & 0.302 & 3.006 & 1.312 & 1.140 & 0.537 & 0.888 & 0.649 & 5.629 & 0.807 & 0.400 & 0.534 & 0.206 & 0.408 \\
Sundial & 0.106 & 0.247 & 0.273 & 0.391 & 0.058 & 0.171 & 0.213 & 0.331 & 2.780 & 1.244 & 1.478 & 0.513 & 1.155 & 0.700 & 6.105 & 0.682 & 0.362 & 0.378 & 0.201 & 0.324 \\
Chronos-2 & 0.103 & 0.245 & 0.257 & 0.380 & 0.060 & 0.179 & 0.144 & 0.252 & 2.291 & 1.096 & 1.186 & 0.523 & 0.798 & 0.607 & 4.786 & 0.673 & 0.301 & 0.358 & 0.185 & 0.309 \\
OFA & 0.100 & 0.242 & \underline{0.247} & 0.378 & 0.057 & 0.180 & \underline{0.129} & 0.245 & 2.708 & 1.250 & 1.522 & 0.610 & 1.065 & 0.699 & 5.659 & 0.868 & 0.383 & 0.421 & 0.251 & 0.369 \\
Time-LLM & 0.097 & 0.240 & 0.250 & 0.379 & 0.057 & 0.178 & 0.138 & 0.254 & 2.648 & 1.230 & 1.758 & 0.701 & 1.025 & 0.707 & 6.450 & 0.971 & 0.422 & 0.447 & 0.253 & 0.368 \\
PromptCast & 0.123 & 0.322 & 0.299 & 0.372 & 0.085 & 0.241 & 0.149 & 0.296 & 4.404 & 1.200 & 1.870 & 0.637 & 1.229 & 0.708 & 7.302 & 1.172 & 0.565 & 0.449 & 0.316 & 0.454 \\
TokenCast & 0.121 & 0.277 & 0.385 & 0.472 & \textbf{0.055} & 0.177 & 0.137 & 0.259 & 2.926 & 1.265 & 1.485 & 0.547 & 0.973 & 0.702 & 6.363 & 0.961 & 0.412 & 0.411 & 0.215 & 0.319 \\
S$^2$IP-LLM & 0.119 & 0.277 & 0.290 & 0.424 & \underline{0.056} & 0.176 & \underline{0.129} & 0.249 & 2.638 & 1.184 & 1.591 & 0.566 & 1.315 & 0.722 & 5.765 & 0.888 & 0.388 & 0.391 & 0.237 & 0.339 \\
TimeReasoner & 0.114 & 0.233 & 0.303 & 0.369 & 0.072 & 0.205 & 0.183 & 0.307 & 4.553 & 1.600 & 1.299 & 0.466 & 1.082 & 0.760 & 4.644 & 0.650 & 0.605 & 0.464 & 0.305 & 0.399 \\
TimeSeriesScientist & 0.213 & 0.290 & 0.498 & 0.505 & 0.092 & 0.205 & 0.150 & 0.300 & 5.907 & 1.680 & 2.057 & 0.712 & 1.028 & 0.673 & 5.977 & 1.033 & 0.606 & 0.476 & 0.301 & 0.439 \\
AlphaCast & 0.123 & 0.225 & 0.279 & 0.372 & 0.075 & 0.185 & 0.131 & 0.271 & 2.375 & \textbf{0.856} & 1.484 & 0.510 & 1.043 & 0.662 & 5.318 & 0.740 & 0.316 & 0.365 & 0.176 & 0.292 \\
\midrule
\rowcolor[gray]{0.95}
\textbf{\model{}-Zero} & \underline{0.080} & \underline{0.212} & 0.268 & \underline{0.361} & \textbf{0.055} & \underline{0.168} & \textbf{0.127} & \textbf{0.237} & \textbf{1.596} & \underline{0.886} & \underline{0.960} & \underline{0.386} & \underline{0.393} & \underline{0.408} & \textbf{3.939} & \textbf{0.501} & \underline{0.221} & \underline{0.277} & \underline{0.140} & \underline{0.268} \\
\rowcolor[gray]{0.95}
\textbf{\model{}-R1} & \textbf{0.077} & \textbf{0.210} & \textbf{0.242} & \textbf{0.351} & \textbf{0.055} & \underline{0.168} & \textbf{0.127} & \textbf{0.237} & \underline{1.757} & 0.918 & \textbf{0.927} & \textbf{0.375} & \textbf{0.386} & \textbf{0.405} & \underline{4.032} & \underline{0.502} & \textbf{0.172} & \textbf{0.247} & \textbf{0.133} & \textbf{0.260} \\
\bottomrule
\end{tabularx}
\end{table*}

\section{Experiments}
\label{sec:experiments}

We examine whether diagnostic-guided numerical anchoring and contextual revision improve forecasting, then assess workflow variants and the two-stage coordinator training. A final case study illustrates how numerical priors are preserved and revised.

\subsection{Experimental Settings}
\label{sec:experimental_setup}

\subsubsection{Datasets}
\label{sec:evaluation_protocol}
Table~\ref{tab:dataset} summarizes the benchmark datasets used in our evaluation. ETT and Wind are used for long-term forecasting, while EPF focuses on short-term forecasting across multiple electricity markets. ETT includes hourly and sub-hourly observations, Wind provides weather-related covariates, and EPF contains market-specific load and generation covariates.
We adopt chronological splits of 60\%, 20\%, and 20\% for training, validation, and testing on ETT, and 70\%, 10\%, and 20\% on Wind, respectively. For long-term evaluation, the look-back length, horizon, and stride are all 96 steps. Following TimeSeriesScientist~\cite{zhao2025timeseriesscientist}, we evaluate a subset of forecast windows to control LLM inference costs. For EPF, we retain the final 604 days per market, using a 168-hour look-back, 24-hour horizon, and 48-hour stride. Each market has 60 test windows covering 1,440 target hours, omitting intervening days. Unlike the original EPF holdout with daily forecast origins, this protocol evaluates all methods on identical sampled windows. Detailed descriptions of the datasets are provided in Appendix~\ref{app:datasets}.
\begin{table}[t]
\centering
\caption{Statistics of the evaluated datasets. }
\label{tab:dataset}
\footnotesize
\setlength{\tabcolsep}{2pt}
\begin{tabular*}{\columnwidth}{@{\extracolsep{\fill}}lccrcc@{}}
    \toprule
    \textbf{Settings} & \textbf{Dataset} & \textbf{Domain} & \textbf{Length} & \textbf{Variables} & \textbf{Frequency} \\
    \midrule
    \multirow{3}{*}{Long-term}  & ETTh1 \& ETTh2          & Electricity   & 17,420 & 7  & 1 hour \\
                                & ETTm1 \& ETTm2         & Electricity   & 69,680 & 7  & 15 mins \\
                                & Wind  & Energy        & 48,673 & 7 & 15 mins \\
    \midrule
    \multirow{5}{*}{Short-term} & BE            & Energy        & 14,496 & 3  & 1 hour \\
                                & DE            & Energy        & 14,496 & 3 & 1 hour \\
                                & FR            & Energy        & 14,496 & 3  & 1 hour \\
                                & NP            & Energy        & 14,496 & 3  & 1 hour \\
                                & PJM           & Energy        & 14,496 & 3  & 1 hour \\
    \bottomrule
  \end{tabular*}
\end{table}

\subsubsection{Baselines}
\label{sec:baselines}
We compare CastFSR with representative methods from five forecasting families. The statistical baselines are ARIMA \cite{box2015time} and Prophet \cite{taylor2018forecasting}. Neural baselines include DLinear \cite{zeng2023transformers}, ConvTimeNet \cite{cheng2025convtimenet}, PatchTST \cite{Yuqietal-2023-PatchTST}, iTransformer \cite{liuitransformer}, and TimeXer \cite{wang2024timexer}. We also compare with foundation models TimesFM \cite{das2024decoder}, Sundial \cite{liu2025sundial}, and Chronos-2 \cite{ansari2025chronos}. The LLM-based baselines are OFA \cite{zhou2023one}, Time-LLM \cite{jintime}, TokenCast \cite{tao2025values}, S$^2$IP-LLM \cite{pan2024s}, TimeReasoner \cite{cheng2025can}, and PromptCast \cite{xue2023promptcast}. The agentic baselines are TimeSeriesScientist \cite{zhao2025timeseriesscientist} and AlphaCast \cite{zhang2025alphacast}. The comparison covers numerical, language-based, and agentic forecasting approaches.

\begin{table*}[t]
\centering
\caption{Ablation of fast--slow--reflect modules. Comparison of CastFSR-Zero with reported variants labelled w/o fast-thinking, w/o slow reasoning, and w/o reflect across ten datasets.}
\label{tab:module_ablation}
\footnotesize
\setlength{\tabcolsep}{0.8pt}
\renewcommand{\arraystretch}{1.12}
\begin{tabularx}{\textwidth}{@{}l*{20}{>{\raggedleft\arraybackslash}X}@{}}
\toprule
\multirow{3}{*}{\textbf{Variants}} & \multicolumn{10}{c}{Long-term forecasting} & \multicolumn{10}{c}{Short-term forecasting} \\
\cmidrule(lr){2-11} \cmidrule(lr){12-21}
 & \multicolumn{2}{c}{ETTh1} & \multicolumn{2}{c}{ETTh2} & \multicolumn{2}{c}{ETTm1} & \multicolumn{2}{c}{ETTm2} & \multicolumn{2}{c}{Wind} & \multicolumn{2}{c}{BE} & \multicolumn{2}{c}{DE} & \multicolumn{2}{c}{FR} & \multicolumn{2}{c}{NP} & \multicolumn{2}{c}{PJM} \\
\cmidrule(lr){2-3} \cmidrule(lr){4-5} \cmidrule(lr){6-7} \cmidrule(lr){8-9} \cmidrule(lr){10-11} \cmidrule(lr){12-13} \cmidrule(lr){14-15} \cmidrule(lr){16-17} \cmidrule(lr){18-19} \cmidrule(lr){20-21}
 & MSE & MAE & MSE & MAE & MSE & MAE & MSE & MAE & MSE & MAE & MSE & MAE & MSE & MAE & MSE & MAE & MSE & MAE & MSE & MAE \\
\midrule
w/o Fast-Thinking & 0.089 & 0.224 & 0.339 & 0.403 & \underline{0.061} & 0.184 & 0.176 & 0.283 & 2.403 & 1.111 & 1.225 & 0.455 & 0.481 & 0.451 & 4.251 & 0.534 & 0.334 & 0.336 & 0.168 & 0.300 \\
w/o Slow Deliberative Reasoning & \underline{0.086} & \underline{0.216} & \underline{0.298} & \underline{0.373} & \textbf{0.055} & \underline{0.169} & 0.157 & 0.265 & 2.310 & 1.054 & 1.147 & 0.431 & 0.438 & 0.426 & \textbf{3.936} & \underline{0.504} & 0.292 & \underline{0.301} & \underline{0.149} & \underline{0.273} \\
w/o Reflective Evaluation & \underline{0.086} & \underline{0.216} & 0.320 & 0.388 & \textbf{0.055} & \underline{0.169} & \underline{0.139} & \underline{0.255} & \underline{2.032} & \underline{0.997} & \underline{1.011} & \underline{0.398} & \underline{0.421} & \underline{0.416} & 3.952 & 0.505 & \underline{0.284} & 0.306 & 0.150 & 0.274 \\
\midrule
\rowcolor[gray]{0.95}
\textbf{\model{}-Zero} & \textbf{0.080} & \textbf{0.212} & \textbf{0.268} & \textbf{0.361} & \textbf{0.055} & \textbf{0.168} & \textbf{0.127} & \textbf{0.237} & \textbf{1.596} & \textbf{0.886} & \textbf{0.960} & \textbf{0.386} & \textbf{0.393} & \textbf{0.408} & \underline{3.939} & \textbf{0.501} & \textbf{0.221} & \textbf{0.277} & \textbf{0.140} & \textbf{0.268} \\
\bottomrule
\end{tabularx}
\end{table*}

\begin{figure*}[t]
    \includegraphics[width=\textwidth]{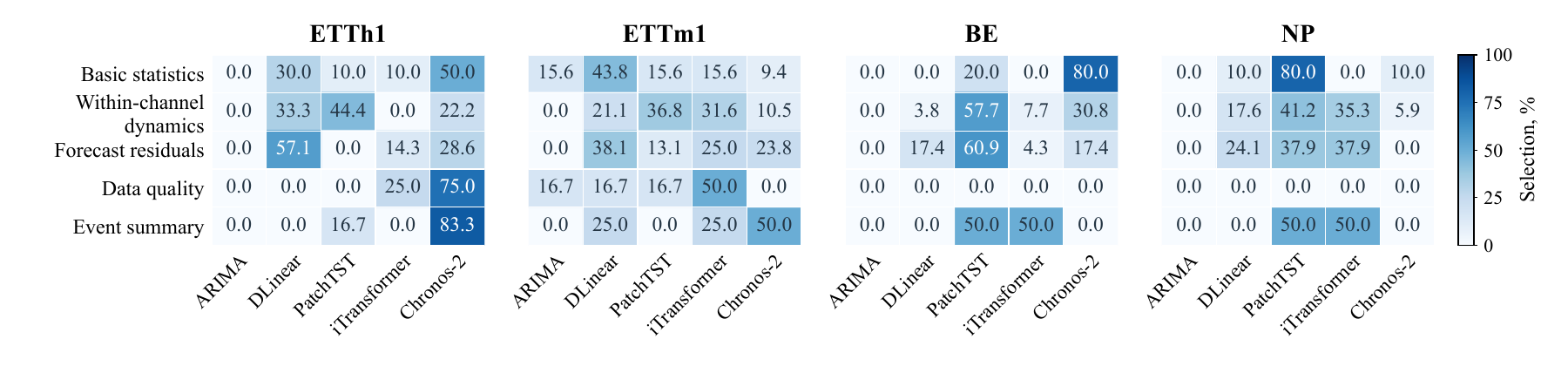}
    \caption{Diagnostic-conditioned routing in \model{}-Zero. Percentages are within diagnostic categories; all-zero rows denote unselected diagnostic categories.}
    \Description{CastFSR-Zero diagnostic-conditioned model-selection frequencies for ETTh1, ETTm1, BE, and NP. Rows indicate diagnostic categories and columns indicate candidate models. The all-zero data quality rows for BE and NP indicate that this diagnostic category was not selected.}
    \label{fig:model_selection_heatmap}
\end{figure*}
\subsubsection{Implementation Details}
\label{sec:implementation}
\label{sec:evaluation_boundaries}
For training-free inference, we instantiate CastFSR-Zero with DeepSeek-V4-Flash~\cite{xu2026deepseek} as the reasoning engine. We further develop CastFSR-R1 using Qwen3-4B~\cite{yang2025qwen3}, initialized through SFT on forecasting trajectories distilled from CastFSR-Zero and further optimized with GRPO using 16 Ascend NPUs. Both variants employ ARIMA, DLinear, PatchTST, iTransformer, and Chronos-2 as numerical forecasting tools, retaining their respective fitting or pretraining procedures. Deep learning baselines follow their official configurations. We report MSE and MAE on the normalized target scale. To prevent information leakage, policy inputs are restricted to historical target observations and covariates, excluding forecast-horizon target and covariate values.
Z-score standardization is applied only for evaluation: final forecasts and ground-truth targets are transformed using the training-partition mean and standard deviation, kept fixed for validation and testing.
See Appendix~\ref{app:implementation_details} for implementation details.

\subsection{Overall Performance Comparison}
\label{sec:main_results}
As shown in Table~\ref{tab:main_results}, we compare CastFSR with numerical and LLM-based forecasting baselines across long- and short-term benchmarks. CastFSR-R1 achieves the best or near-optimal performance on most datasets, demonstrating the effectiveness of the fast--slow--reflect workflow for context-aware forecasting. Several key observations can be made. First, CastFSR achieves competitive performance across diverse forecasting tasks, highlighting the benefits of combining numerical priors with contextual reasoning and reflective evaluation to capture contextual influences on future dynamics. Second, CastFSR-Zero performs competitively without task-specific coordinator training, showing that the workflow can be effectively implemented with off-the-shelf LLMs. Third, CastFSR-R1 further improves forecasting accuracy on most datasets, suggesting the benefits of adapting compact LLMs through post-training. However, CastFSR-Zero performs better on Wind and FR, indicating that coordinator effectiveness varies across forecasting tasks. Overall, these results support the effectiveness of integrating fast-thinking forecasting, slow deliberative reasoning, and reflective evaluation for context-aware forecasting.

\subsection{Analysis of the Fast--Slow--Reflect Workflow}
\label{sec:workflow_analysis}

\subsubsection{Ablation Study of the Workflow Modules}
\label{sec:workflow_ablation}
\label{sec:reflective_consistency}

Table~\ref{tab:module_ablation} evaluates each component by removing fast-thinking forecasting, slow deliberative reasoning, or reflective evaluation from CastFSR-Zero. Removing individual components generally degrades performance, confirming their complementary roles. Removing fast-thinking forecasting causes the largest degradation, indicating that data-driven priors provide a critical foundation for subsequent reasoning. Without these priors, the agent relies more on LLM generation, leading to less accurate forecasts. Removing slow deliberative reasoning also reduces performance on most metrics, highlighting the importance of contextual reasoning. Similarly, removing reflective evaluation increases forecasting errors on most metrics, demonstrating its value in checking temporal and contextual consistency.

\subsubsection{Analysis of Forecaster Selection}
\label{sec:expert_preferences}
\label{sec:forecaster_selection}

Figure~\ref{fig:model_selection_heatmap} illustrates the forecaster selection patterns of \model{}-Zero during fast-thinking forecasting. The selection distributions vary across diagnostic categories and datasets, indicating that the coordinator adapts its forecaster choices to different temporal characteristics rather than relying on a fixed model. These observations suggest that diagnostic evidence guides numerical forecaster selection to construct data-driven forecast priors.
Table~\ref{tab:model_selection_gain} compares adaptively selected experts with the best fixed expert in the candidate pool for each dataset, using the same expert as the reference for both metrics. The comparison considers direct expert forecasts before slow or reflect. Adaptive selection improves forecasting accuracy in several cases, although fixed experts remain more effective in others. These results highlight the potential benefits of evidence-guided forecaster selection while revealing its dataset-dependent effectiveness.

\begin{table}[!t]
\centering
\caption{Comparison of \model{}-Zero selection and the lowest-MSE fixed expert in the candidate pool.}
\footnotesize
\setlength{\tabcolsep}{1pt}
\renewcommand{\arraystretch}{1.1}
\begin{tabular*}{\columnwidth}{@{\extracolsep{\fill}}l|rr|rr|rr|rr@{}}
\toprule
\textbf{Strategies} & \multicolumn{2}{c|}{ETTh1} & \multicolumn{2}{c|}{ETTm1} & \multicolumn{2}{c|}{BE} & \multicolumn{2}{c}{NP} \\
& MSE & MAE & MSE & MAE & MSE & MAE & MSE & MAE \\
\midrule
Best Model Result & 0.091 & \textbf{0.213} & 0.060 & 0.179 & 1.164 & 0.495 & \textbf{0.301} & \textbf{0.358} \\
\textbf{\model{} selection} & \textbf{0.088} & 0.225 & \textbf{0.057} & \textbf{0.177} & \textbf{1.098} & \textbf{0.448} & 0.337 & 0.374 \\
\midrule
Best Model & \multicolumn{2}{c|}{iTransformer} & \multicolumn{2}{c|}{Chronos-2} & \multicolumn{2}{c|}{PatchTST} & \multicolumn{2}{c}{Chronos-2} \\
\bottomrule
\end{tabular*}
\captionsetup{position=bottom}

\label{tab:model_selection_gain}
\end{table}

\subsubsection{Analyzing Adaptive Slow Deliberative Reasoning}
\label{sec:adaptive_slow_reasoning}

\begin{figure}
    \includegraphics[width=\linewidth]{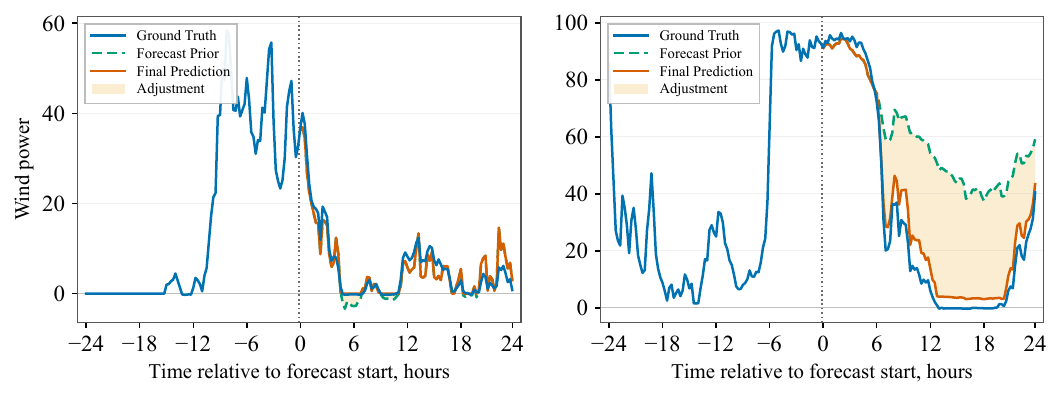}
    \caption{Two wind-power windows: history $t<0$, forecasts $t\ge0$; shading marks prior-to-final change.}
    \Description{Two wind-power examples comparing numerical priors and final complete-workflow predictions with observed targets.}
    \label{fig:context_adjustment}
\end{figure}

Figure~\ref{fig:context_adjustment} illustrates how CastFSR selectively refines numerical forecast priors in two wind-power examples. Slow deliberative reasoning assesses historical contextual evidence to determine whether revisions are warranted, preserving useful prior structure while introducing targeted adjustments when needed. Shaded regions highlight the timing and magnitude of these changes. The displayed curves reflect the combined effects of slow deliberative reasoning and reflective evaluation, without separately showing intermediate predictions. Across all datasets, 66.71\% of forecasts remain essentially unchanged, while 19.92\% and 13.37\% undergo mild and substantial revisions, respectively. These findings highlight CastFSR's tendency to preserve numerical priors rather than uniformly replacing them.

\begin{figure}
    \includegraphics[width=\linewidth]{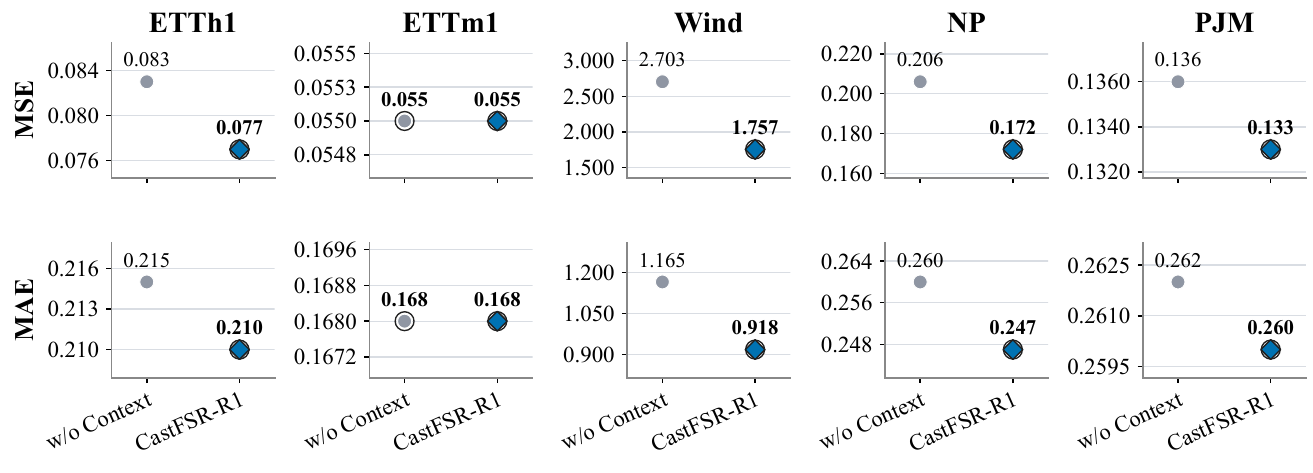}
    \caption{CastFSR-R1 context ablation in slow with a shared numerical prior. Panels use independent scales.}
    \Description{Dot plots comparing CastFSR-R1 with the variant labelled w/o context on ETTh1, ETTm1, Wind, NP, and PJM. Only historical contextual evidence in the slow stage is removed; both variants retain the same numerical prior.}
    \label{fig:context_ablation}
\end{figure}

\subsubsection{Analysis of Context Utilization}
\label{sec:context_refinement}

Figure~\ref{fig:context_ablation} examines the contribution of historical contextual evidence to slow deliberative reasoning in CastFSR-R1. Both variants share the same numerical prior, with historical covariates retained in the numerical expert's inputs. The w/o context variant removes contextual evidence only from the slow stage, isolating its contribution to forecast refinement. Incorporating this evidence generally improves forecasting accuracy, although the gains vary across scenarios. These findings highlight the additional value of explicit contextual reasoning beyond information already incorporated in the numerical prior.

\subsection{Analysis of Agent Instantiations}
\label{sec:instantiation_analysis}

\subsubsection{Effect of Training Strategies}
\label{sec:training_strategies}

\begin{figure}
    \includegraphics[width=\linewidth]{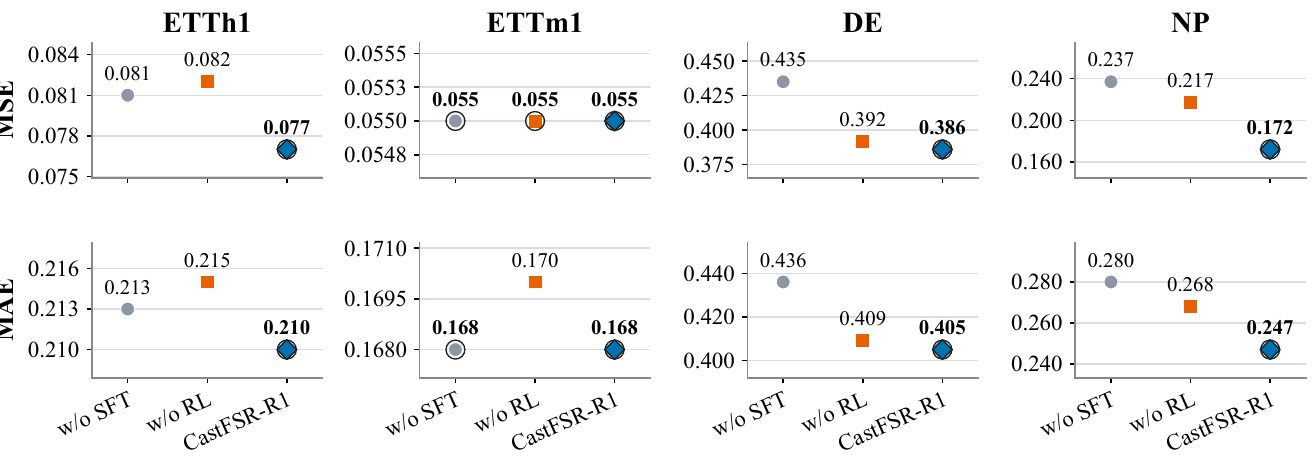}
    \caption{Impact of SFT and RL on forecasting performance. }
    \Description{Dot plots comparing CastFSR-R1 with variants trained without SFT or without RL on MSE and MAE.}
    \label{fig:training_stage_ablation}
\end{figure}

Figure~\ref{fig:training_stage_ablation} investigates the contributions of SFT and RL by comparing CastFSR-R1 with variants that remove either stage. The complete model achieves comparable performance, demonstrating the effectiveness of combining supervised trajectory distillation and reinforcement learning. Removing SFT leads to substantial performance degradation, particularly on DE and NP, highlighting the importance of learning structured forecasting decisions from teacher trajectories. Removing RL reduces forecasting accuracy, indicating that policy optimization helps improve decisions beyond supervised learning. The contributions of these stages vary across datasets, suggesting their complementary roles in adapting the coordinator to diverse temporal dynamics. These findings support the effectiveness of the WATD and TLPO training framework in improving context-aware performance. 
\subsubsection{Impact of Pre-training LLMs}
\label{sec:llm_coordinators}
Table~\ref{tab:training_free_api} compares \model{}-Zero using different pretrained LLM coordinators. The framework achieves competitive forecasting performance without task-specific training, although results vary across backbones and datasets. Notably, training-free Qwen3-4B outperforms its post-trained counterpart on one dataset, suggesting that post-training may introduce trade-offs in dataset-specific performance. Nevertheless, \model{}-R1 achieves substantial improvements over training-free Qwen3-4B on the other evaluated datasets.

\subsection{Case Study Analysis}
\label{sec:case_study}

\begin{figure*}[t]
\includegraphics[width=\linewidth]{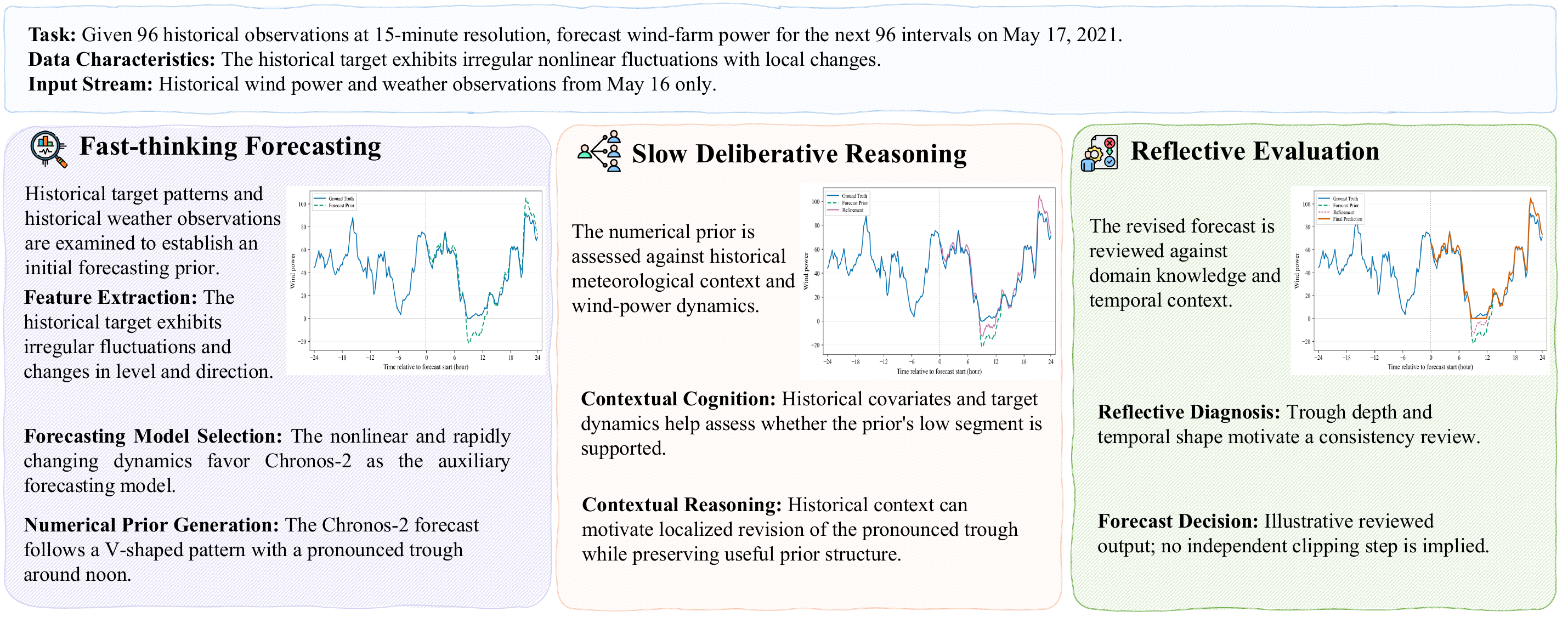}
\caption{
Illustrative case study of \model{} in context-aware time series forecasting. }
\Description{An illustrative wind-power forecasting walkthrough showing numerical-prior generation, contextual refinement, and reflective assessment.}
\label{fig:case_study}
\end{figure*}

Figure~\ref{fig:case_study} shows a wind-power walkthrough of CastFSR. During fast-thinking forecasting, the coordinator extracts temporal patterns and constructs a data-driven forecast prior using a selected numerical expert. The prior contains a pronounced trough, motivating an assessment against historical wind and target evidence. During slow deliberative reasoning, contextual evidence guides localized revisions while useful prior structure is retained. Reflective evaluation then checks the candidate against temporal, contextual, and domain rules, feeding detected violations back for correction. The stage-specific annotations are schematic, illustrating the coordination of numerical priors, contextual reasoning, and~forecast~review. 
\begin{table}[t]
\centering
\caption{Training-free forecasting performance of CastFSR-Zero with different pretrained LLMs.}
\label{tab:training_free_api}

\footnotesize
\setlength{\tabcolsep}{1.5pt}
\renewcommand{\arraystretch}{1.15}
\begin{tabular*}{\columnwidth}{@{\extracolsep{\fill}}l*{8}{r}@{}}
\toprule
\multirow{2}{*}{\textbf{LLM Coordinator}}
& \multicolumn{2}{c}{ETTh1}
& \multicolumn{2}{c}{ETTm1}
& \multicolumn{2}{c}{DE}
& \multicolumn{2}{c}{NP} \\
\cmidrule(lr){2-3} \cmidrule(lr){4-5} \cmidrule(lr){6-7} \cmidrule(lr){8-9}
& MSE & MAE
& MSE & MAE
& MSE & MAE
& MSE & MAE \\
\midrule
GPT-5.6-sol
& \underline{0.081} & 0.213
& 0.055 & 0.168
& 0.400 & 0.414
& 0.218 & \underline{0.266} \\

LongCat-2.0
& 0.089 & 0.219
& 0.056 & 0.168
& \textbf{0.383} & \underline{0.412}
& 0.221 & 0.271 \\

DeepSeek-V4-Pro
& \underline{0.081} & \underline{0.212}
& \underline{0.054} & \underline{0.165}
& 0.403 & 0.417
& \underline{0.216} & 0.267 \\

DeepSeek-V4-Flash
& \textbf{0.080} & \underline{0.212}
& 0.055 & 0.168
& \underline{0.393} & \textbf{0.408}
& 0.221 & 0.277 \\

GLM-5.2
& \textbf{0.080} & \textbf{0.210}
& 0.055 & 0.167
& 0.405 & 0.427
& \textbf{0.207} & \textbf{0.260} \\

Qwen3-4B
& 0.086 & 0.217
& \textbf{0.050} & \textbf{0.161}
& 0.452 & 0.435
& 0.324 & 0.369 \\
\bottomrule
\end{tabular*}
\end{table}

\section{Conclusion and Discussion}
In this paper, we proposed \model{}, a fast--slow--reflect agentic framework for context-aware time series forecasting. The framework integrates numerical prior construction, evidence-guided refinement, and reflective review through a shared state that retains evidence and forecasting rationale. Fast-thinking forecasting establishes a data-driven prior, while slow reasoning selectively refines it using contextual evidence. Reflective evaluation reviews the revised forecast against temporal patterns and domain constraints, assessing whether revisions are justified and guiding iterative correction. The framework supports both training-free coordination with general-purpose LLMs and a compact 4B coordinator trained to learn the fast--slow--reflect policy. WATD and TLPO combine state-conditioned trajectory supervision with reinforcement learning guided by forecasting accuracy and workflow feedback. Experiments on ten public energy forecasting datasets under an offline protocol demonstrate competitive performance.

\paragraph{\textbf{Discussion}.} CastFSR highlights an important insight: contextual information does not always justify revising numerical predictions. Reliable forecasting requires preserving useful numerical patterns while making selective, evidence-supported revisions. Despite its structured workflow, several limitations remain. First, reliance on historical context limits its ability to account for unexpected future events or newly available evidence. Second, LLM-based reflective evaluation cannot guarantee satisfaction of all domain constraints. Moreover, multi-turn reasoning and tool interactions introduce inference overhead that warrants further evaluation. Future work will explore richer contextual evidence, more reliable constraint checking, and more efficient agentic forecasting workflows.

\clearpage
\bibliographystyle{ACM-Reference-Format}
\bibliography{references}

\appendix
\section{Detailed Dataset Descriptions}
\label{app:datasets}

\subsection{Long-term Time Series Forecasting}

\begin{itemize}
    \item \textbf{ETT:} In the electricity transformer temperature benchmark \cite{zhou2021informer}, oil temperature (OT) is the forecasting target, and HUFL, HULL, MUFL, MULL, LUFL, and LULL are the six load features. ETTh1 and ETTm1 measure one transformer, while ETTh2 and ETTm2 measure the other. All four series start at 2016-07-01 00:00; the hourly series end at 2018-06-26 19:00 and the 15-minute series at 2018-06-26 19:45, inclusive.
    \item \textbf{Wind:} This dataset is distributed with the TFB benchmark \cite{qiu2024tfb} and covers 2020-01-01 00:00 through 2021-05-22 00:00, inclusive. Wind power is the target. Its contextual fields are predicted humidity, pressure, temperature, wind direction, wind speed, and observed wind speed.
\end{itemize}

\subsection{Short-term Electricity Price Forecasting}

The EPF data come from Lago et al.~\cite{lago2021forecasting}, with market price as the target and ComEd prices used for PJM. Covariates are day-ahead forecasts: BE uses French load and generation; DE uses aggregated wind and solar generation and Amprion load; FR uses load and generation; NP uses grid load and wind generation; and PJM uses system and zonal loads.
For each of BE, DE, FR, NP, and PJM, we retain the final 14,496 hourly observations, or 604 days, to accommodate SFT dataset construction and general-purpose LLM inference costs. The same truncation rule is applied separately to each market. The training, validation, and test partitions contain 10,224, 1,416, and 2,856 hours, respectively, with the following exact boundaries. Each range begins at 00:00 on its first day and ends at 23:00 on its last day, inclusive:
\begin{itemize}
    \item \textbf{BE and FR:} training 2015-05-08--2016-07-06; validation 2016-07-07--2016-09-03; testing 2016-09-04--2016-12-31.
    \item \textbf{DE:} training 2016-05-07--2017-07-06; validation 2017-07-07--2017-09-03; testing 2017-09-04--2017-12-31.
    \item \textbf{NP and PJM:} training 2017-04-30--2018-06-29; validation 2018-06-30--2018-08-27; testing 2018-08-28--2018-12-24.
\end{itemize}
The retained segment extends from the first training timestamp to the last test timestamp. Validation and test examples reuse the preceding 168 hours as historical context; their forecast targets remain within the assigned partition.

\subsection{Test Windows and Sampling}
\label{app:test_windows}

Table~\ref{tab:test_windows} reports the evaluated time ranges and window counts. The range starts with the first test example's historical input and ends with the last example's forecast target; it does not imply that every timestamp in the range is scored. Test windows are sampled deterministically at these strides, without random selection. Each forecast origin is the first target timestamp. The first origin is 2018-02-01 00:00 for all four ETT datasets and 2021-02-10 00:00 for Wind. For EPF, the first origins are 2016-09-04 00:00 for BE and FR, 2017-09-04 00:00 for DE, and 2018-08-28 00:00 for NP and PJM.
Each EPF market has 60 origins spaced 48 hours apart. Each window scores the next 24 hourly values, evaluating 60 days and omitting the intervening 59 days. All methods use the same forecast origins and target timestamps under this protocol. The EPF comparisons therefore apply to these selected segments and sampled origins, rather than the original holdout under standard daily evaluation.
\begin{table}[!htbp]
\centering
\caption{Test coverage and window counts, from the first look-back window to the last scored target, inclusive.}
\label{tab:test_windows}
\footnotesize
\setlength{\tabcolsep}{3pt}
\renewcommand{\arraystretch}{1.12}
\begin{tabular*}{\columnwidth}{@{\extracolsep{\fill}}llr@{}}
\toprule
\textbf{Datasets} & \textbf{Start--End} & \textbf{Windows} \\
\midrule
ETTh1 & 2018-01-28 00:00 -- 2018-06-24 23:00 & 36 \\
ETTh2 & 2018-01-28 00:00 -- 2018-06-24 23:00 & 36 \\
ETTm1 & 2018-01-31 00:00 -- 2018-06-25 23:45 & 145 \\
ETTm2 & 2018-01-31 00:00 -- 2018-06-25 23:45 & 145 \\
Wind & 2021-02-09 00:00 -- 2021-05-21 23:45 & 101 \\
\midrule
NP & 2018-08-21 00:00 -- 2018-12-24 23:00 & 60 \\
PJM & 2018-08-21 00:00 -- 2018-12-24 23:00 & 60 \\
BE & 2016-08-28 00:00 -- 2016-12-31 23:00 & 60 \\
DE & 2017-08-28 00:00 -- 2017-12-31 23:00 & 60 \\
FR & 2016-08-28 00:00 -- 2016-12-31 23:00 & 60 \\
\bottomrule

\end{tabular*}
\end{table}

\section{Detailed Baseline Descriptions}
\label{app:baselines}

We compare \model{} with 18 baselines spanning five forecasting families.
\textbf{ARIMA} \cite{box2015time} models temporal dependencies through autoregression, differencing, and moving averages. \textbf{Prophet} \cite{taylor2018forecasting} uses additive components for trend, seasonality, and holidays.
\textbf{DLinear} \cite{zeng2023transformers} applies linear prediction to decomposed trend and seasonal components. \textbf{ConvTimeNet} \cite{cheng2025convtimenet} uses hierarchical convolutions to capture multi-scale temporal patterns. \textbf{PatchTST} \cite{Yuqietal-2023-PatchTST} combines patch tokenization with channel-independent transformers. \textbf{iTransformer} \cite{liuitransformer} applies attention across variate-level tokens. \textbf{TimeXer} \cite{wang2024timexer} integrates endogenous patches and exogenous variables through attention.
\textbf{TimesFM} \cite{das2024decoder} is a pretrained decoder-only model for zero-shot forecasting. \textbf{Sundial} \cite{liu2025sundial} provides pretrained foundation models for zero-shot and transfer forecasting. \textbf{Chronos-2} \cite{ansari2025chronos} supports zero-shot forecasting with multivariate inputs and covariates.
\textbf{OFA (one fits all)} \cite{zhou2023one} adapts pretrained language models with frozen backbone layers. \textbf{Time-LLM} \cite{jintime} reprograms numerical inputs for LLM-based forecasting. \textbf{TokenCast} \cite{tao2025values} represents numerical values as symbolic tokens for context-aware forecasting. \textbf{S$^2$IP-LLM} \cite{pan2024s} uses semantic-informed prompt learning. \textbf{TimeReasoner} \cite{cheng2025can} explores slow-thinking reasoning for forecasting. \textbf{PromptCast} \cite{xue2023promptcast} formulates forecasting as a language prompting task.
\textbf{TimeSeriesScientist} \cite{zhao2025timeseriesscientist} employs an LLM-driven agent for time series analysis. \textbf{AlphaCast} \cite{zhang2025alphacast} integrates human knowledge and LLM reasoning for interactive forecasting.

\section{Detailed Implementation Settings}

\label{app:implementation_details}

\subsection{CastFSR-Zero}
\label{app:coordinator_configuration}

CastFSR-Zero uses DeepSeek-V4-Flash as the default coordinator without task-specific parameter updates. The coordinator acquires diagnostic evidence, selects a numerical expert, refines the returned prior, and reviews the candidate forecast through the fast--slow--reflect workflow. The released configuration specifies temperature 0.6, top-$p$ 0.7, and a maximum output length of 32,768 tokens. Tool-stage generation defaults to 512 tokens. Training-free refers to the coordinator; the numerical experts retain their own fitting or pretraining procedures.

\subsection{CastFSR-R1}

CastFSR-R1 uses Qwen3-4B as the coordinator. SFT imitates forecasting trajectories generated by CastFSR-Zero, followed by GRPO optimization of multi-turn interactions. The SFT and GRPO learning rates are $2.0 \times 10^{-6}$ and $2.0 \times 10^{-7}$, respectively. Training uses 16 Ascend NPUs. The RL configuration specifies global batch size 128, group size $G=5$, generation temperature 1.0, and maximum prompt/response lengths of 20,480/6,144 tokens. We generate candidate multi-turn forecasting trajectories using CastFSR-Zero on the training data. We first filter out incomplete trajectories and invalid forecasting outputs, then select valid trajectories based on forecasting quality. Filtering yields 2,584 SFT trajectories. Training comprises 646 SFT steps followed by 80 RL steps.

\begin{figure*}[!t]
    \centering
    \includegraphics[width=0.98\textwidth]{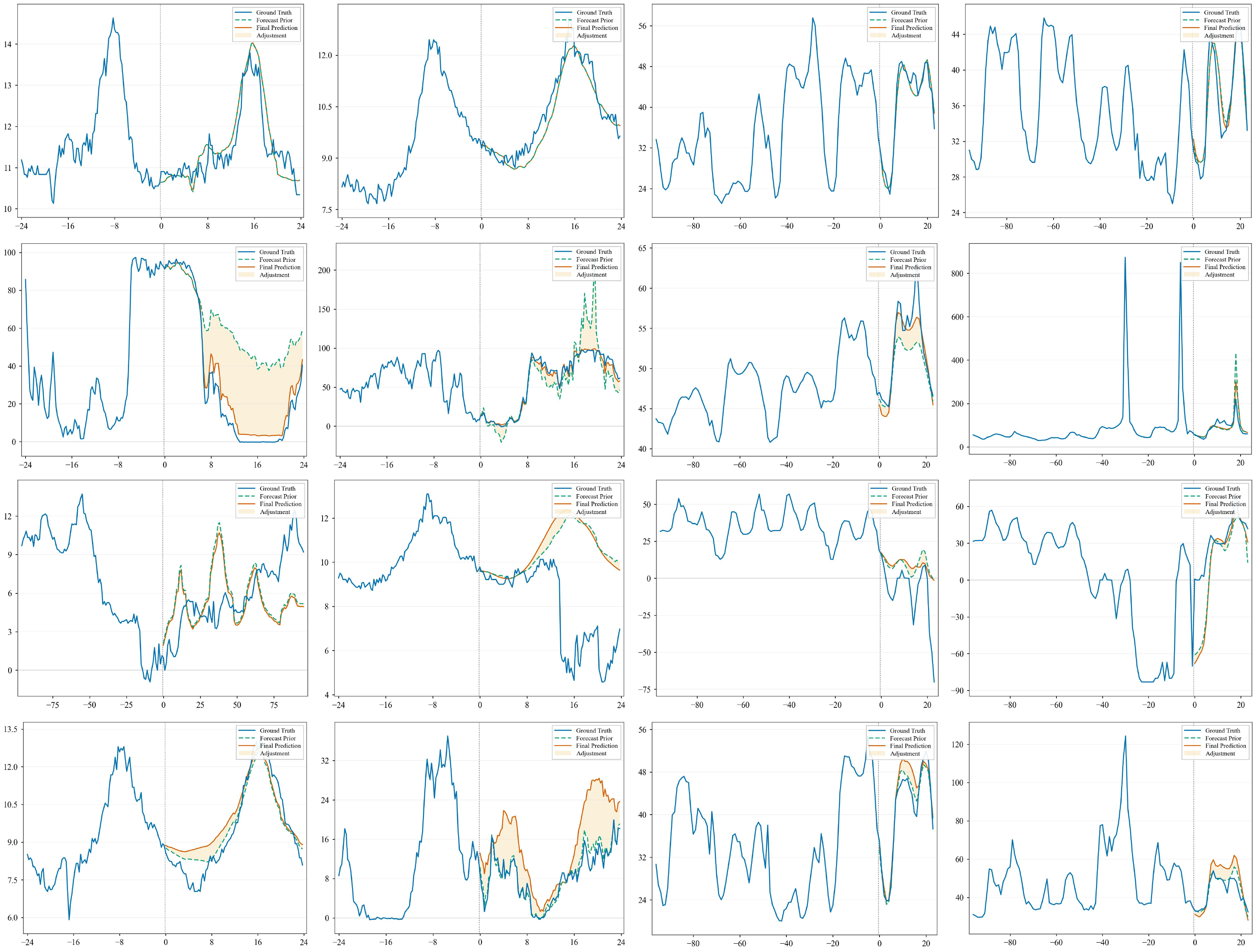}
    \caption{Visualization of successful and unsuccessful forecast refinements across sixteen representative examples.  The upper rows show successful refinements, while the lower rows highlight failures.}
    \Description{Sixteen time-series panels compare ground truth, numerical forecast priors, and final predictions around a marked forecast origin. Shading highlights the difference between each prior and final prediction. Examples include accurate tracking, corrected declines and spikes, missed changes, and excessive adjustments.}
    \label{fig:context_window_case}
\end{figure*}
\subsection{Forecasting Experts and Tool Configuration}
\label{sec:execution}

The shared expert pool comprises ARIMA, DLinear, PatchTST, iTransformer, and Chronos-2. The predictor service refits ARIMA to each request's target history with default order $(2,1,2)$. Neural experts cache dataset-specific configurations and checkpoints and run in evaluation mode without online gradient updates. The evaluated expert calls use historical target observations and historical covariates according to their input interfaces.
The default preference is ARIMA, the first predictor in the released order; no validation-based selection procedure is recorded. Offset/scale calibration defaults to \texttt{none} and requires explicit configuration. Malformed optional guard decisions use the deterministic recommended route. The released Wind guard bounds are $[-2,102]$; applying these rules to normalized targets requires a verified scale transform.
A predictor retry retains diagnostic evidence but clears the prior; a restart clears diagnostic caches and analysis history. Both rebuild prompts. Limits are five LLM-loop iterations, five tool calls per iteration, and three optional prior-guard rounds. Reflect permits up to three correction retries, passing detected rule violations to the next prompt. When needed, one retry typically sufficed in observed runs; reflect's retry budget was never exhausted. Valid answers terminate execution. Exhaustion without an accepted answer constitutes policy failure.  Forecasts are essentially unchanged if all prior-to-final absolute differences are at most 0.0005. Otherwise, revision RMSE normalized by the training-target standard deviation separates mild and substantial revisions at 0.10. 

\begin{ForecastPrompt}[label=app:agent_prompt,label=pro:reflection_prompt]
\textbf{Role and Inputs.}
Forecast the next $H$ values using historical targets and covariates, task metadata, domain priors, and tool evidence. Use original units throughout; exclude forecast-horizon target and covariate values.

\textbf{Stage 1---Fast-thinking Forecasting.}
Select one diagnostic view: statistics, dynamics, residuals, data quality, or events. Use its evidence to choose one expert: ARIMA, DLinear, PatchTST, iTransformer, or Chronos-2. Obtain a numerical prior; retain diagnostic evidence and expert identity.

\textbf{Stage 2---Slow Deliberative Reasoning.}
Summarize the situation and refine the prior using historical context. Preserve temporal alignment and useful structure; revise only where evidence supports the direction and timing. Pass the candidate to reflect.

\textbf{Stage 3---Reflective Evaluation.}
Check temporal, contextual, and domain rules. Feed violations into the next prompt for correction and recheck, with at most three retries. Retain supported structure; keep these checks separate from format, finite-value, horizon-length, and timestamp checks.

\textbf{Output.}
Give a brief situation analysis and slow rationale in \texttt{<think>}. Return exactly $H$ timestamp--finite-decimal rows in one \texttt{<answer>} block:
\begin{verbatim}
<think>Situation and forecast reasoning</think>
<answer>YYYY-MM-DD HH:MM:SS finite_decimal
...
YYYY-MM-DD HH:MM:SS finite_decimal</answer>
\end{verbatim}
\end{ForecastPrompt}

\subsection{Prompt Construction and Output Format}
\label{app:prompt_construction}

Template~\ref{app:agent_prompt} summarizes the stage-specific prompt instructions. At runtime, each turn combines a system prompt with a structured state snapshot containing task metadata, historical target observations, historical contextual covariates, and accumulated tool evidence. The situation and covariate block is cached within a run; retrieval follows historical observation timestamps. Slow and reflect execute in separate generations, with violation feedback supplied on correction retries. 

\section{Reward Definition and Training Credit}
\label{app:reward}

The reward mechanism combines forecasting accuracy, workflow progress, and output validity. Forecasting accuracy is evaluated using MSE and MAE with a base weight of 0.50 and an optional expert-relative improvement weight of 0.10, with the accuracy component clipped to $[-0.60, 0.60]$. Valid output format, forecast length, and timestamp alignment receive rewards of 0.10, 0.15, and 0.15, respectively, while successful intermediate tool progress receives 0.02. Protocol violations incur penalties for malformed outputs ($-0.75$), stalled responses ($-0.60$), out-of-stage actions ($-0.65$), and truncated outputs ($-0.70$). The terminal reward is clipped to $[-1,1]$, with protocol failures capped at $-0.25$ and unsuccessful completion receiving a final-step penalty of at most $-0.75$.
Step rewards form a trajectory return, with group-normalized advantages.

\section{Additional Experimental Results}
\label{app:additional_results}

\subsection{Visualization Analysis}
\label{app:visualization_analysis}

Figure~\ref{fig:context_window_case}  presents representative cases of successful and unsuccessful forecast refinements. In successful cases, the final predictions better capture temporal variations and local patterns, demonstrating the benefits of refining numerical priors. However, some adjustments introduce additional errors or fail to capture abrupt changes, highlighting the limitations of contextual reasoning. These observations suggest that forecast refinement can improve accuracy but does not consistently outperform the original priors. 

\subsection{Inference Efficiency}
\label{app:inference_efficiency}

Table~\ref{tab:inference_efficiency} compares the inference efficiency of CastFSR. Slow deliberative reasoning dominates inference time, while CastFSR-R1 incurs higher latency and token consumption than CastFSR-Zero. Trajectory analysis reveals distinct recovery behaviors, with CastFSR-R1 favoring workflow restarts and CastFSR-Zero adopting more localized retries. Under the evaluated deployments, CastFSR-Zero also achieves higher generation throughput, reflecting differences in recovery strategies and serving configurations. Overall, CastFSR-Zero offers flexible, training-free coordination with efficient API serving but relies on external services, whereas CastFSR-R1 enables task-specific adaptation and self-hosted deployment, offering greater control over sensitive data

\begin{table}[t]
\centering

\caption{Inference efficiency of CastFSR.}
\label{tab:inference_efficiency}
\footnotesize
\setlength{\tabcolsep}{2pt}
\renewcommand{\arraystretch}{1.12}
\begin{tabular*}{\columnwidth}{@{\extracolsep{\fill}}lrrrrrr@{}}
\toprule
\textbf{Methods} & \makecell[r]{\textbf{Fast}\\(s) $\downarrow$} & \makecell[r]{\textbf{Slow}\\(s) $\downarrow$} & \makecell[r]{\textbf{Reflect}\\(s) $\downarrow$} & \makecell[r]{\textbf{Total}\\(s) $\downarrow$} & \makecell[r]{\textbf{Tokens}\\(K) $\downarrow$} & \makecell[r]{\textbf{LLM}\\\textbf{calls} $\downarrow$} \\
\midrule
CastFSR-Zero & 0.989 & 58.410 & 3.877 & 63.276 & 54.821 & 5 \\
CastFSR-R1 & 4.788 & 66.571 & 5.112 & 76.469 & 65.930 & 6 \\
\bottomrule
\end{tabular*}
\end{table}

\end{document}